\documentclass{article}

\PassOptionsToPackage{square,sort,comma,numbers}{natbib}

    \usepackage[final]{neurips_2020_ml4ps}

\usepackage[utf8]{inputenc} 
\usepackage[T1]{fontenc}    
\usepackage{hyperref}       
\usepackage{url}            
\usepackage{booktabs}       
\usepackage{amsfonts}       
\usepackage{nicefrac}       
\usepackage{microtype}      
\usepackage{xcolor}
\usepackage{amsmath}
\usepackage{graphicx}

\usepackage{fontawesome}
\newcommand{\github}{\href{https://github.com/NiallJeffrey/MomentNetworks}{\faGithub}}

\title{Solving high-dimensional parameter inference: marginal posterior densities \& Moment Networks}

\author{%
Niall Jeffrey $^{1,2}$\\
$^{1}$ Laboratoire de Physique de l'{\'E}cole Normale Sup\'erieure, \\ ENS, Universit\'e PSL, CNRS, Sorbonne Universit\'e, Universit\'e de Paris, Paris, France \\ 
$^{2}$ Department of Physics \& Astronomy, University College London, Gower Street, London, UK \\ 
\texttt{niall.jeffrey@phys.ens.fr} \\
\And
Benjamin D.~Wandelt $^{3,4}$ \\
$^{3}$Institut d'Astrophysique de Paris (IAP), UMR 7095, CNRS, Sorbonne Universit\'e, France\\
$^{4}$Center for Computational Astrophysics, Flatiron Institute, 162 5th Avenue, New York, USA\\
   \texttt{bwandelt@iap.fr}
}

\begin{document}

\maketitle

\begin{abstract}
  High-dimensional probability density estimation for inference suffers from the ``curse of dimensionality''. For many physical inference problems, the full posterior distribution is unwieldy and seldom used in practice. Instead, we propose direct estimation of lower-dimensional marginal distributions, bypassing high-dimensional density estimation or high-dimensional Markov chain Monte Carlo (MCMC) sampling. By evaluating the two-dimensional marginal posteriors we can unveil the full-dimensional parameter covariance structure. We additionally propose constructing a simple hierarchy of fast neural regression models, called Moment Networks, that compute increasing moments of any desired lower-dimensional marginal posterior density; these reproduce exact results from analytic posteriors and those obtained from Masked Autoregressive Flows. We demonstrate marginal posterior density estimation using high-dimensional LIGO-like gravitational wave time series and describe applications for problems of fundamental cosmology. \github
\end{abstract}



\section{Introduction} 
\vspace{-0.3cm}
Estimating the posterior probability density $ p(\boldsymbol{\theta} | \boldsymbol{x} )$  of a set of parameters $\boldsymbol{\theta} $  given some observed data $\boldsymbol{x}$ is often the primary objective of problems of inference, prediction, or generation. The object $ p(\boldsymbol{\theta} | \boldsymbol{x} )$ encapsulates all belief and uncertainties about the unknown quantities $\boldsymbol{\theta}$. With this aim in mind, recent advances in neural density estimation have improved our ability to estimate the density $ p(\boldsymbol{\theta} | \boldsymbol{x})$ from a set of training examples $\{ \boldsymbol{x}_i,  \boldsymbol{\theta}_i \}$.

Estimating such probability densities with neural density methods, such as Mixture Density Networks \cite{mdn}, or recent state-of-the-art normalizing flow methods, such as Masked Autogregressive Flows (MAF  \cite{MAF}), provide an excellent way to quantify uncertainty for predicted or inferred parameters and signals $\boldsymbol{\theta}$. Used for likelihood-free inference (also known as simulation-based inference \cite{Brehmer5242,Cranmer201912789}) these density estimation methods can estimate conditional probability densities for parameters and data, either the posterior or the likelihood ~\cite{Papamakarios_lfi, delfi2}. 

For high-dimensional signals, estimation of the full joint density is often not useful and, instead, summaries of lower-dimensional marginal densities are the final goal. For example, the marginal posterior density per pixel, or subsets of pixels, could serve to quantify uncertainty in a reconstructed image.

In this example, the joint posterior marginal for pairs of pixel parameters $\boldsymbol{\theta} = [ \alpha, \beta ]$ given some observed data $\boldsymbol{x}_{obs}$
\begin{equation} \label{eq:marg}
p(\alpha, \beta | \boldsymbol{x}_{obs} ) = \int p(\alpha, \beta,  \boldsymbol{\theta}' | \boldsymbol{x}_{obs} ) \  \mathrm{d}  \boldsymbol{\theta}'\ \ ,
\end{equation}
\noindent would marginalize over all possible values of all other parameters (i.e. the other pixels and latent parameters)  $\boldsymbol{\theta}'$. If this were evaluated for all pairs of parameters, all 2D marginal moments of the high-dimensional posterior distribution would be characterized.

In this contribution we present two complementary approaches to evaluate the two-dimensional marginal posterior distributions, marginal flows and Moment Networks (Sec.~\ref{sec:density_estimation}). In Sec.~\ref{sec:experiments} we demonstrate the two methods in comparison to a known underlying posterior density (sampled with MCMC), and show a simulated gravitational wave data model, where the underlying time-ordered signal values form the high-dimensional parameter space to be inferred. In Sec.~\ref{sec:cosmo} we describe seemingly intractable problems in cosmological inference that can be solved using marginal posterior density estimation. 

\section{Marginal posterior density estimation} \label{sec:density_estimation} 
\vspace{-0.1cm}
\vspace{-0.15cm} \paragraph{Motivation} In practice, for many physical inference problems with high-dimensional parameter spaces, the full high-dimensional posterior distribution $p(\boldsymbol{\theta} | \boldsymbol{x}_{obs} ) $ is not necessary or even interpretable. Instead, the inference goal are the marginal one- and two-dimensional posterior distributions of the parameters \cite[e.g.][] {planck18,kids18,des_year1}.

Even if full posterior sampling is possible through sophisticated MCMC techniques, e.g. in hierarchical models with known distributions (rather than in a likelihood-free framework), the number of posterior samples needed to compute a $d$-dimensional marginal grows exponentially with  $d$. For high-dimensional problems, the limited number of independent samples that result  in practice  only allow for the computation of low-dimensional marginals of the posterior density.

We therefore make the marginal densities the target of our inference problem. We take inspiration from the simplicity of integration when a  Monte Carlo sampled representation of the posterior distribution is available. In this case, marginalization is trivial: it amounts simply to ignoring the parameter dimensions to be marginalized over. In this work we show that we can directly bring this powerful notion to  simulation-based inference; it allows us to estimate the marginal posterior density (or its moments)  directly. This is a  powerful approach whenever we are dealing with a large number parameters and effectively removes the practical limitation of simulation-based inference techniques to applications with a low-dimensional parameter space.
\begin{figure*}
\vspace{-0.8cm}
\centering
\includegraphics[width=1.0\textwidth]{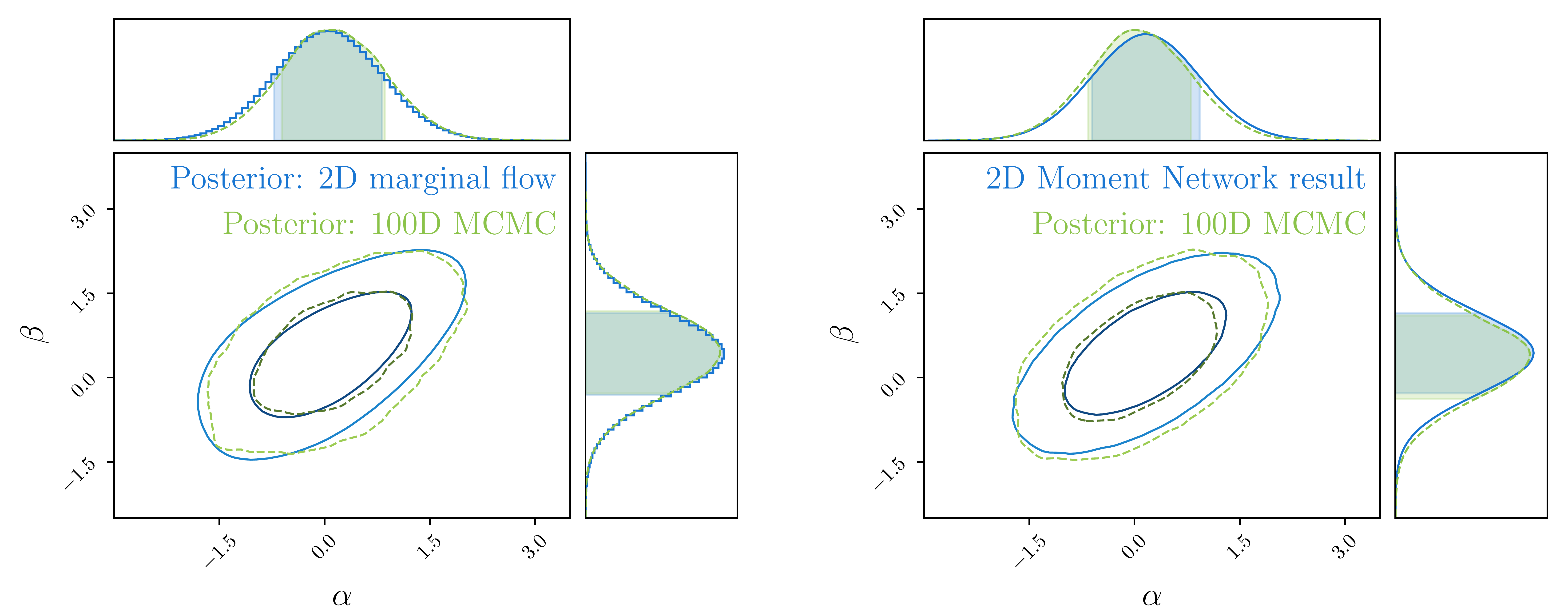}
\vspace{-0.3cm}
\caption{100-dimensional data model with known reference distribution evaluated with $10^7$ MCMC samples. Direct 2D marginal posterior estimation using a MAF ensemble (\textit{left panel}) and representation of 2D Moment Network result (\textit{right panel}) both trained with {\color{black} $8\times10^4$ simulations.}}
\vspace{-0.5cm}
\label{fig:joint_marginal}
\end{figure*}
\vspace{-0.15cm} \paragraph{Marginal flows} Many popular and powerful density estimation methods can be categorized as \textit{normalizing flows}. These use a series of bijective functions to transform from simple known densities (e.g. unit normal) to the target density \cite{normalizing_flows, kingma2016improved}. MAFs represent the estimated density $q$ as a transformation of a unit normal through a series of autoregressive functions \cite{MAF,MAF2}. The networks are trained to give an estimate $q$ of the target distribution $p$ by minimizing a Monte Carlo estimate of the Kullback-Leibler divergence~\cite{kullback1951}. For a sampling distribution $p(\boldsymbol{x} | \boldsymbol{\theta})$ this would be
\begin{equation} \label{eq:delfi_loss}
U(\boldsymbol{\varphi}) = - \sum_{i=1}^N \log q  (\boldsymbol{x}_i |  \boldsymbol{\theta}_i ; \boldsymbol{\varphi}) \ \ 
\end{equation}
\noindent with varying network parameters $\boldsymbol{\varphi}$ over the forward-modelled mock data $\boldsymbol{x}_i$. In this same likelihood-free framework, one can directly estimate the posterior distributions for subsets of the large parameter set. For any two parameters $\alpha$ and $\beta$ of the full $\boldsymbol{\theta}$, one can directly estimate $p(\alpha,\beta | \boldsymbol{x})$ by minimizing
\begin{equation} \label{eq:delfi_loss2}
U(\boldsymbol{\varphi}) = - \sum_{i=1}^N \log q  (\alpha_i, \beta_i | \boldsymbol{x}_i ; \boldsymbol{\varphi}) \ \ . 
\end{equation}
The resulting density will indeed be an estimate of the marginal posterior for the chosen parameter pair (eq.~\ref{eq:marg}) if all parameters of $\boldsymbol{\theta}$ (not just the chosen pair) are drawn from the prior $p(\boldsymbol{\theta})$ to generate the training data. This procedure also avoids the need for data compression steps~\cite{imnn,alsing_compression}, as we condition on high-dimensional data $\boldsymbol{x}$ rather than estimating its density, and any nuisance parameters are automatically marginalized away, provided they have been sufficiently sampled in the training data. 
\begin{figure*}[b]
\vspace{-0.4cm}
\centering
\includegraphics[width=.95\textwidth]{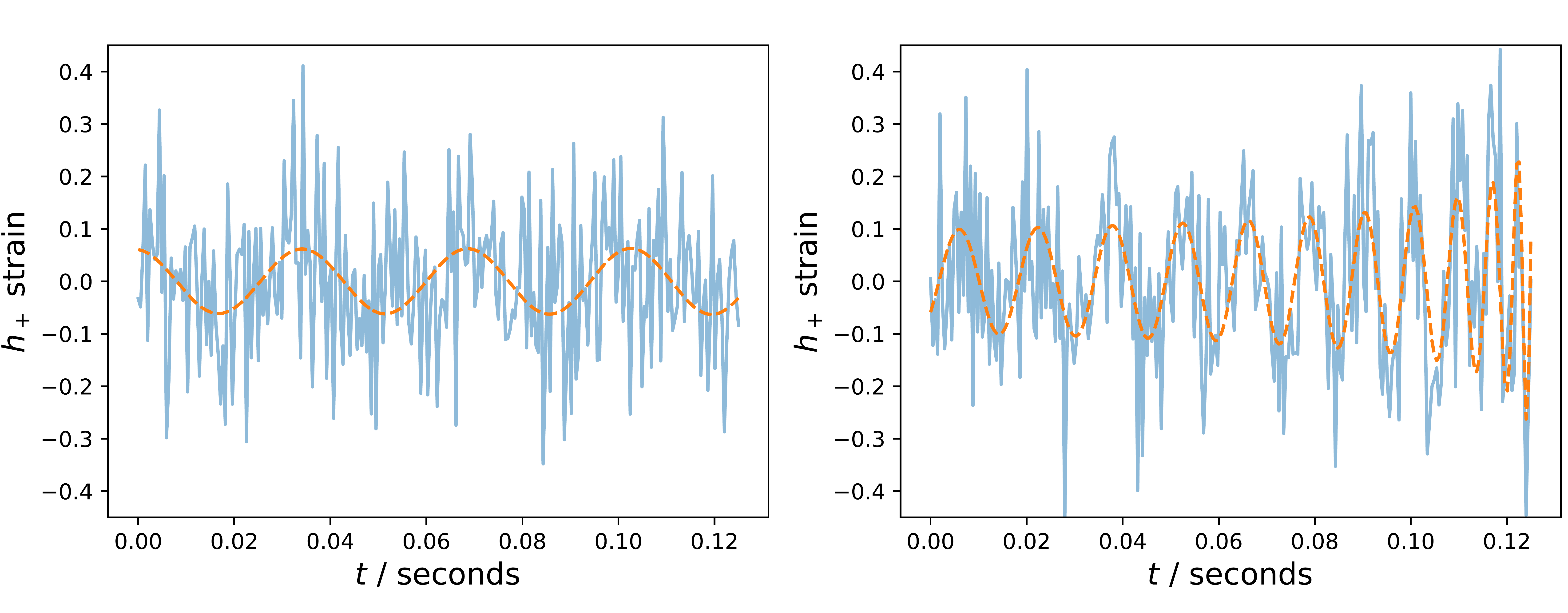}
\vspace{-0.4cm}
\caption{Two example simulated gravitation wave time series signals for the strain ``$+$'' polarization $h_{+}$ with realistic LIGO-like noise. The dashed line shows the true strain values over time.}
\vspace{-0.5cm}
\label{fig:strain}
\end{figure*}

\vspace{-0.15cm} \paragraph{Moment Networks} In practice, posterior estimates often serve principally to compute posterior moments.  Moment Networks allow us to side-step the problem of estimating the posterior density and directly skip to estimation of location, scale, and covariance of the parameters (and possibly higher-order moments). When this is sufficient, Moment Networks allow the use of far simpler neural network architectures, which reduces risk of training failure, and boosts inference speed. 

We begin by noting that if we find some function of our data $\mathcal{F}(\boldsymbol{x})$ that minimizes an $L_2$ loss over the distribution of possible training examples $\{ \boldsymbol{x}_i,  \boldsymbol{\theta}_i \}$,
\begin{equation}
    J_0 = \int || \boldsymbol{\theta} - \mathcal{F}(\boldsymbol{x}) ||^2 p(\boldsymbol{x},  \boldsymbol{\theta}  ) \ \mathrm{d}\boldsymbol{x} \  \mathrm{d}\boldsymbol{\theta} \ \ ,
\end{equation}
\noindent then $\mathcal{F}$, which we represent as a neural network, evaluated for the observed data is the mean of the posterior distribution $\mathcal{F}(\boldsymbol{x}_{obs}) = \langle \boldsymbol{\theta} \rangle_{\boldsymbol{\theta} | \boldsymbol{x}_{obs}}$.
It is therefore possible to create a hierarchy of networks to generate further moments of the posterior distribution. For example, the function $\mathcal{G}$ that minimizes
\begin{equation}
    J_1 = \int || (\boldsymbol{\theta} - \mathcal{F}_{\mathrm{fixed}}(\boldsymbol{x}))^2 - \mathcal{G}(\boldsymbol{x}) ||^2 p(\boldsymbol{x},  \boldsymbol{\theta}) \  \mathrm{d}\boldsymbol{x} \  \mathrm{d}\boldsymbol{\theta} \ \ ,
\end{equation}
for fixed, already trained $\mathcal{F}$, is such that $\mathcal{G}(\boldsymbol{x}_{obs})$, is the set of posterior variances ~\cite{jaynes07,adler}. The objective functions for marginal posterior parameter covariances can be similarly constructed.

By sampling the full parameter space from the prior distributions $p(\boldsymbol{\theta})$, the functions $\mathcal{F}$ and $\mathcal{G}$ can be combined to output the posterior means, variances, and covariances for subsets of the full set of parameters; the marginalization over other parameters is implicitly done during training. This result is exact and independent of the true underlying posterior or prior distributions. 

{\color{black} The Moment Network solves for the marginal posterior moments by construction, and therefore does not suffer the problem of mode collapse in variational inference with multi-modal posteriors, which can lead to underpredicted uncertainty. Outside the likelihood-free framework, if one does have information about the functional form of the posterior, one can fit the posterior parameters to the marginal moments (see Sec.~\ref{sec:cosmo}).}

\section{Experiments} \label{sec:experiments}
\textbf{High-dimensional
inference}: We use a 100-dimensional parameter inference toy model to demonstrate marginal posterior estimation for pairs of parameters. The model consists of 100-element data vectors with non-stationary Gaussian noise and a Gaussian prior distribution with non-trivial covariance introducing parameter correlation. We estimate the marginal posterior density for parameter pairs using MAF (using the \texttt{pyDELFI} package \cite{delfi2}) and estimate the 2D marginal mean, variance and covariance using Moment Networks. The results are represented in Fig.~\ref{fig:joint_marginal}.

As a reference, we can directly sample the posterior distribution using high-dimensional MCMC, which took $10^7$ draws from the likelihood. The normalizing flow result would have been intractable if the density estimation target was with respect to the full parameter space or to the data space. With a marginal flow, changing the target to the pairs of parameters, it is simple (with a basic 2-GPU:12GB set-up) to evaluate all marginal posterior pairs (often represented as a so-called ``corner plot'').

With the same set-up, the Moment Network hierarchy was able to accurately evaluate the means, variances and covariances of the marginals (see Fig.~\ref{fig:joint_marginal}) with a few seconds of training and evaluation in $\mathcal{O}(10^{-2})$s without requiring any sampling or grid evaluation. For many practical applications of inference in the physical sciences, these marginal joint moments would be the final goal.

\textbf{Gravitational wave signal demonstration}: Fig.~\ref{fig:strain} shows two example simulated gravitational wave time series. The two signals (\textit{dashed orange}) are $\sim0.12$s intervals from the $1$s before a binary black hole merger using the \texttt{SEOBNRv4} model \cite{SEOBNRv4_opt}.

We have simplified the problem for this demonstration by removing all ``geometric'' effects (black hole spin, inclination, detector geometry) and use only the $+$ polarization for the detector strain $h_{+,\times}$. We do, however, sample the merger events with an independent prior distribution for each mass $p(M)=\mathcal{U}(10,30) M_{\mathrm{sol}}$ and distance $p(\chi)=\mathcal{U}(500,1500) \mathrm{Mpc}$. The noise is LIGO-like noise, which, along with the signal, was generated using the \texttt{pyCBC} package for 35000 simulations.

\begin{figure}[t]
\centering
\includegraphics[width=1.0\textwidth]{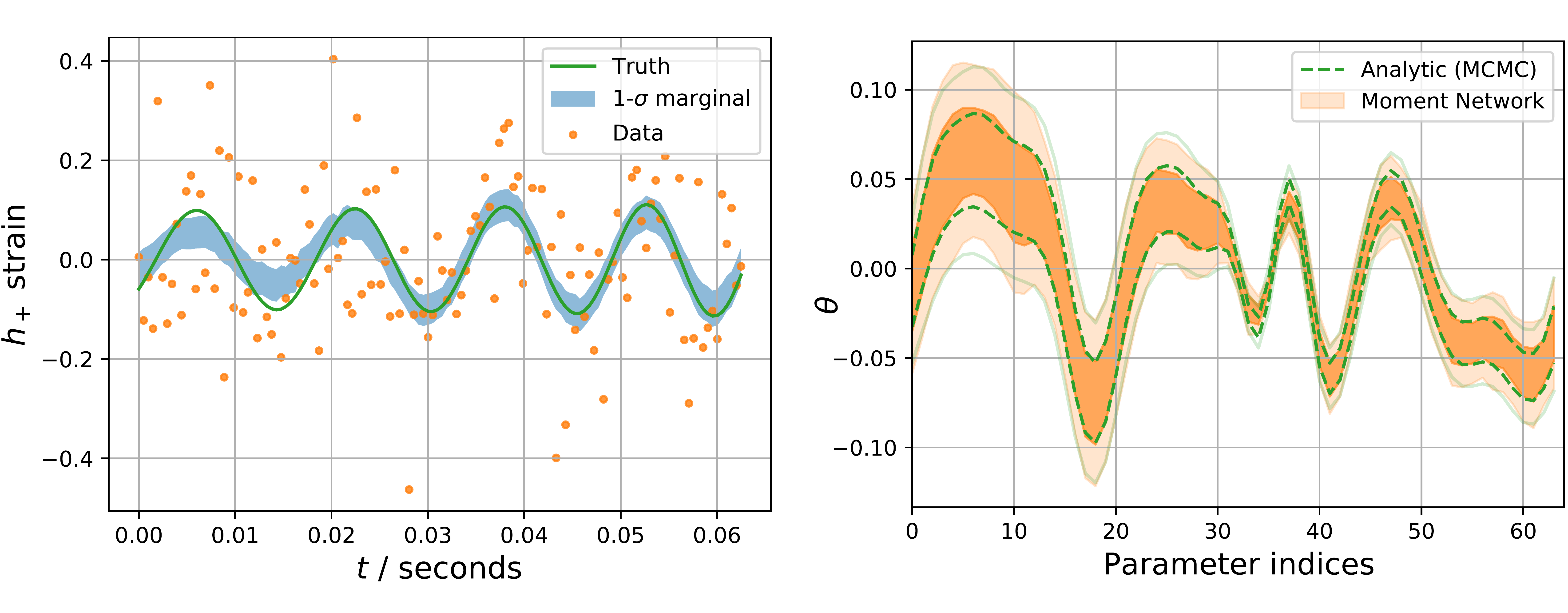}
\caption{\textit{Left panel}: Moment Network (MN) estimate of the 1-$\sigma$ standard deviation per strain $h_+(\Delta t)$ parameter given 62.5ms of data (Fig.~\ref{fig:strain}). \textit{Right panel}: For each of 64 parameters (c.f. time step), contours are the marginal posterior $\sigma$ from MN (\textit{shaded orange}) and MCMC (\textit{dashed green}).}
\label{fig:posterior_strain}
\end{figure}

With the time series elements forming a high-dimensional parameter space, the \textit{left panel} of Fig.~\ref{fig:posterior_strain} shows a representation of the marginal posterior standard deviation for each of 128 parameters for a simulated data set. This result was evaluated using the trained Moment Network. The \textit{right panel} shows a validation case of similar complexity to the gravitational wave model, but with known likelihood. The Moment Network trained on simulations matches accurately with a long-run MCMC chain. This validates our approach.

\newpage

\section{Discussion \& cosmological applications} \label{sec:cosmo}
Though the direct density estimation of marginal posteriors is much more robust than the estimation of the full posterior, it may still suffer from well-known issues of density estimation. Moment Networks optimize a completely different set of objective functions to return estimates of the posterior moments. This affords an opportunity to cross-validate, as moments of the estimated marginal posteriors should match those from the Moment Network. If the results are inconsistent for an initial set of simulations, then there may be insufficient network complexity (the network complexity for both methods scales similarly) or insufficient number of simulations.

Thus far, density estimation likelihood-free inference in cosmology has generally been limited to a few parameters \cite[e.g.][]{delfi2, Taylor_2019, Brehmer_2019, des_lfi, ramanah2020dynamical,2020arXiv201008537L}. Though simulation-based inference of cosmological fields (including dark matter) can be integral to many analyses \cite[e.g.][]{Caldeira_2019, deeplearning_shirasaki, deepmass,petroff2020fullsky}, it can be intractable to estimate the full posterior due to high-dimensionality. With the approach we propose, joint marginal posteriors (and associated moments) for reconstructed cosmological fields can be directly evaluated.

For cases where it is possible to sample, marginal flows and Moment Networks still provide advantages. One particularly ambitious cosmological sampler  BORG~\cite{Jasche_2013, Jasche_2015} samples the $\sim10^7\mathrm{-}$dimensional posterior density of the initial conditions of the Universe for given galaxy data, using a non-linear forward model including the physics and data effects. Though the full posterior is sampled, the complexity of the sampler and the inherently sequential nature of MCMC limits the number of independent samples to $\mathcal{O}(10^3)$; sufficient only to estimate low-dimensional marginal posteriors and their moments. The approach proposed in this work could use similar computational resources to generate simulations to train marginal flows and Moment Networks in parallel (rather than sequentially) and efficiently output low-dimensional marginal posteriors and moments.

Beyond general high-dimensional inference, the principal motivation for this work (Sec.~\ref{sec:density_estimation}),  we plan to explore a wide range of further applications of marginal flows and Moment Networks to probe the fundamental physics of the Universe in future studies.

\hspace{1cm}

Demonstration code can be found at: {\color{blue} \tt \small \href{https://github.com/NiallJeffrey/MomentNetworks}{github.com/NiallJeffrey/MomentNetworks}}








\section*{Broader Impact}

\vspace{-0.1cm}

This work provides a robust approach to quantify uncertainty from high-dimensional parameter spaces by estimating marginal posterior distributions or their associated moments directly. This has immediate application for parameter and model inference in astrophysics and cosmology, and the physical sciences more generally. We note that if the method is misunderstood or misapplied, incorrect uncertainty quantification or risk analysis would follow. {\color{black} To mitigate this, diagnostic and validation methods can be applied (e.g. ensembles of neural density estimators or quantile tests) or, as proposed in this work, by comparing results between likelihood-free methods (e.g. marginal flows and Moment Networks).} The approach in this work can be applied to signal inference and prediction more generally (including fast image analysis, time series prediction, forecasting, and quantifying uncertainty for decision making).

\section*{Acknowledgements}

\vspace{-0.1cm}
\paragraph{Software used} \texttt{pyDELFI} ({\url{github.com/justinalsing/pydelfi}}) for MAF density estimation implementation~\cite{delfi2}; \texttt{chainconsumer} ({\url{samreay.github.io}}) for Fig.~\ref{fig:joint_marginal} \cite{Hinton2016}; \texttt{emcee} ({emcee.readthedocs.io}) for MCMC sampling \cite{emcee}; \texttt{pyCBC} ({\url{pycbc.org}}) for gravitational wave data simulation~\cite{pycbc1}.

{\color{black} The authors thank Tom Charnock for useful discussions. NJ acknowledges funding from the {\'E}cole Normale Sup{\'e}rieure (ENS). BDW acknowledges support by the ANR BIG4
project, grant ANR-16-CE23-0002 of the French Agence Nationale
de la Recherche; and the Labex ILP (reference ANR-10-LABX-63)
part of the Idex SUPER, and received financial state aid managed by the Agence Nationale de la Recherche, as part of the programme Investissements d’avenir under the reference ANR-11-IDEX-0004-02. The Flatiron Institute is supported by the Simons Foundation.}



\bibliographystyle{abbrv}
\bibliography{neurips_2020}

\begin{thebibliography}{10}

\bibitem{des_year1}
T.~M.~C. {Abbott}, F.~B. {Abdalla}, A.~{Alarcon}, J.~{Aleksi{\'c}}, S.~{Allam},
  S.~{Allen}, A.~{Amara}, J.~{Annis}, J.~{Asorey}, S.~{Avila}, and et~al.
\newblock {Dark Energy Survey year 1 results: Cosmological constraints from
  galaxy clustering and weak lensing}.
\newblock {\em PRD}, 98(4):043526, Aug 2018.

\bibitem{adler}
J.~{Adler} and O.~{{\"O}ktem}.
\newblock {Deep Bayesian Inversion}.
\newblock {\em arXiv e-prints}, page arXiv:1811.05910, Nov. 2018.

\bibitem{delfi2}
J.~{Alsing}, T.~{Charnock}, S.~{Feeney}, and B.~{Wand elt}.
\newblock {Fast likelihood-free cosmology with neural density estimators and
  active learning}.
\newblock {\em MNRAS}, 488(3):4440--4458, Sept. 2019.

\bibitem{alsing_compression}
J.~{Alsing} and B.~{Wandelt}.
\newblock {Generalized massive optimal data compression}.
\newblock {\em MNRAS}, 476(1):L60--L64, May 2018.

\bibitem{mdn}
C.~Bishop.
\newblock Mixture density networks.
\newblock Working \ paper, Aston University, 1994.

\bibitem{SEOBNRv4_opt}
A.~Bohé, L.~Shao, A.~Taracchini, A.~Buonanno, S.~Babak, I.~W. Harry,
  I.~Hinder, S.~Ossokine, M.~Pürrer, V.~Raymond, and et~al.
\newblock Improved effective-one-body model of spinning, nonprecessing binary
  black holes for the era of gravitational-wave astrophysics with advanced
  detectors.
\newblock {\em Physical Review D}, 95(4), Feb 2017.

\bibitem{Brehmer5242}
J.~Brehmer, G.~Louppe, J.~Pavez, and K.~Cranmer.
\newblock Mining gold from implicit models to improve likelihood-free
  inference.
\newblock {\em Proceedings of the National Academy of Sciences},
  117(10):5242--5249, 2020.

\bibitem{Brehmer_2019}
J.~Brehmer, S.~Mishra-Sharma, J.~Hermans, G.~Louppe, and K.~Cranmer.
\newblock Mining for dark matter substructure: Inferring subhalo population
  properties from strong lenses with machine learning.
\newblock {\em The Astrophysical Journal}, 886(1):49, Nov 2019.

\bibitem{Caldeira_2019}
J.~Caldeira, W.~Wu, B.~Nord, C.~Avestruz, S.~Trivedi, and K.~Story.
\newblock Deepcmb: Lensing reconstruction of the cosmic microwave background
  with deep neural networks.
\newblock {\em Astronomy and Computing}, 28:100307, Jul 2019.

\bibitem{imnn}
T.~{Charnock}, G.~{Lavaux}, and B.~D. {Wandelt}.
\newblock {Automatic physical inference with information maximizing neural
  networks}.
\newblock {\em PRD}, 97(8):083004, Apr. 2018.

\bibitem{Cranmer201912789}
K.~Cranmer, J.~Brehmer, and G.~Louppe.
\newblock The frontier of simulation-based inference.
\newblock {\em Proceedings of the National Academy of Sciences}, 2020.

\bibitem{emcee}
D.~{Foreman-Mackey}, D.~W. {Hogg}, D.~{Lang}, and J.~{Goodman}.
\newblock {emcee: The MCMC Hammer}.
\newblock {\em PASP}, 125(925):306, Mar. 2013.

\bibitem{Hinton2016}
S.~R. {Hinton}.
\newblock {ChainConsumer}.
\newblock {\em The Journal of Open Source Software}, 1:00045, Aug. 2016.

\bibitem{Jasche_2015}
J.~Jasche, F.~Leclercq, and B.~Wandelt.
\newblock Past and present cosmic structure in the sdss dr7 main sample.
\newblock {\em Journal of Cosmology and Astroparticle Physics},
  2015(01):036–036, Jan 2015.

\bibitem{Jasche_2013}
J.~Jasche and B.~D. Wandelt.
\newblock Bayesian physical reconstruction of initial conditions from
  large-scale structure surveys.
\newblock {\em Monthly Notices of the Royal Astronomical Society},
  432(2):894–913, Apr 2013.

\bibitem{jaynes07}
E.~T. Jaynes.
\newblock {\em Probability theory: the logic of science}.
\newblock Cambridge University Press, 2003.

\bibitem{des_lfi}
N.~Jeffrey, J.~Alsing, and F.~Lanusse.
\newblock Likelihood-free inference with neural compression of des sv weak
  lensing map statistics, 2020.

\bibitem{deepmass}
N.~{Jeffrey}, F.~{Lanusse}, O.~{Lahav}, and J.-L. {Starck}.
\newblock {Deep learning dark matter map reconstructions from DES SV weak
  lensing data}.
\newblock {\em MNRAS}, 492(4):5023--5029, Mar. 2020.

\bibitem{normalizing_flows}
D.~{Jimenez Rezende} and S.~{Mohamed}.
\newblock {Variational Inference with Normalizing Flows}.
\newblock {\em arXiv e-prints}, page arXiv:1505.05770, May 2015.

\bibitem{kids18}
S.~{Joudaki} and {KiDS Collaboration}.
\newblock {KiDS-450 + 2dFLenS: Cosmological parameter constraints from weak
  gravitational lensing tomography and overlapping redshift-space galaxy
  clustering}.
\newblock {\em MNRAS}, 474(4):4894--4924, Mar. 2018.

\bibitem{kingma2016improved}
D.~P. Kingma, T.~Salimans, R.~Jozefowicz, X.~Chen, I.~Sutskever, and
  M.~Welling.
\newblock Improved variational inference with inverse autoregressive flow.
\newblock In {\em Advances in neural information processing systems}, pages
  4743--4751, 2016.

\bibitem{kullback1951}
S.~Kullback and R.~A. Leibler.
\newblock On information and sufficiency.
\newblock {\em Ann. Math. Statist.}, 22(1):79--86, 03 1951.

\bibitem{2020arXiv201008537L}
P.~{Lemos}, N.~{Jeffrey}, L.~{Whiteway}, O.~{Lahav}, N.~I. {Libeskind}, and
  Y.~{Hoffman}.
\newblock {The sum of the masses of the Milky Way and M31: a likelihood-free
  inference approach}.
\newblock {\em arXiv e-prints}, page arXiv:2010.08537, Oct. 2020.

\bibitem{pycbc1}
A.~Nitz, I.~Harry, D.~Brown, C.~M. Biwer, J.~Willis, T.~D. Canton, C.~Capano,
  L.~Pekowsky, T.~Dent, A.~R. Williamson, G.~S. Davies, S.~De, M.~Cabero,
  B.~Machenschalk, P.~Kumar, S.~Reyes, D.~Macleod, dfinstad, F.~Pannarale,
  T.~Massinger, M.~Tápai, L.~Singer, S.~Kumar, S.~Khan, S.~Fairhurst,
  A.~Nielsen, SSingh087, shasvath, B.~U.~V. Gadre, and I.~Dorrington.
\newblock gwastro/pycbc: Pycbc release v1.16.10.
\newblock {\em Zenodo}, Oct. 2020.

\bibitem{Papamakarios_lfi}
G.~Papamakarios and I.~Murray.
\newblock Fast $\varepsilon$-free inference of simulation models with bayesian
  conditional density estimation.
\newblock {\em Advances in Neural Information Processing Systems}, pages
  1028--1036, 2016.

\bibitem{MAF}
G.~Papamakarios, T.~Pavlakou, and I.~Murray.
\newblock Masked autoregressive flow for density estimation.
\newblock {\em Advances in Neural Information Processing Systems}, pages
  2338--2347, 2017.

\bibitem{MAF2}
G.~Papamakarios, D.~Sterratt, and I.~Murray.
\newblock Sequential neural likelihood: Fast likelihood-free inference with
  autoregressive flows.
\newblock In K.~Chaudhuri and M.~Sugiyama, editors, {\em Proceedings of Machine
  Learning Research}, volume~89 of {\em Proceedings of Machine Learning
  Research}, pages 837--848. PMLR, 16--18 Apr 2019.

\bibitem{petroff2020fullsky}
M.~A. Petroff, G.~E. Addison, C.~L. Bennett, and J.~L. Weiland.
\newblock Full-sky cosmic microwave background foreground cleaning using
  machine learning, 2020.

\bibitem{planck18}
PlanckCollaboration.
\newblock {Planck 2018 results. VI. Cosmological parameters}.
\newblock {\em AAP}, 641:A6, Sept. 2020.

\bibitem{ramanah2020dynamical}
D.~K. Ramanah, R.~Wojtak, Z.~Ansari, C.~Gall, and J.~Hjorth.
\newblock Dynamical mass inference of galaxy clusters with neural flows, 2020.

\bibitem{deeplearning_shirasaki}
M.~{Shirasaki}, N.~{Yoshida}, and S.~{Ikeda}.
\newblock {Denoising weak lensing mass maps with deep learning}.
\newblock {\em PRD}, 100(4):043527, Aug. 2019.

\bibitem{Taylor_2019}
P.~L. Taylor, T.~D. Kitching, J.~Alsing, B.~D. Wandelt, S.~M. Feeney, and J.~D.
  McEwen.
\newblock Cosmic shear: Inference from forward models.
\newblock {\em Physical Review D}, 100(2), Jul 2019.

\end{thebibliography}





\end{document}